\documentclass[sn-mathphys-num]{sn-jnl}
\usepackage{graphicx}%
\usepackage{multirow}%
\usepackage{amsmath,amssymb,amsfonts}%
\usepackage{amsthm}%
\usepackage{mathrsfs}%
\usepackage[title]{appendix}%
\usepackage{xcolor}%
\usepackage{textcomp}%
\usepackage{manyfoot}%
\usepackage{booktabs}%
\usepackage{algorithm}%
\usepackage{algorithmicx}%
\usepackage{algpseudocode}%
\usepackage{listings}%
\usepackage{array}
\usepackage{booktabs} 
\usepackage{diagbox}
\usepackage{multirow}
\usepackage{booktabs}  
\usepackage{array} 
\usepackage{caption} 
\usepackage{setspace}
\usepackage{natbib}
\usepackage{geometry}
\usepackage[compact]{titlesec} 
\newcolumntype{P}[1]{>{\centering\arraybackslash}p{#1}}
\newcolumntype{M}[1]{>{\centering\arraybackslash}m{#1}}
\usepackage{multirow}  
\usepackage{subcaption}
\usepackage{tikz}
\usetikzlibrary{decorations.pathreplacing}
\usepackage{setspace}  

\begin{document}
\title[Article Title]{The Back Propagation of the Wave Network}

\author[1]{\fnm{Xin} \sur{Zhang}}\email{zha19053@ttu.edu}
\author[2]{\fnm{Victor S.} \sur{Sheng}}\email{victor.sheng@ttu.edu}

\affil[1]{\orgdiv{Department of Computer Science}, \orgname{Texas Tech University}, \orgaddress{\street{2500 Broadway}, \city{Lubbock}, \state{Texas}, \postcode{79409}, \country{USA}}}

\affil[2]{\orgdiv{Department of Computer Science}, \orgname{Texas Tech University}, \orgaddress{\street{2500 Broadway}, \city{Lubbock}, \state{Texas}, \postcode{79409}, \country{USA}}}

\abstract{This paper provides an in-depth analysis of Token2Wave, a novel token representation method derived from the Wave Network, designed to capture both global and local semantics of input text through wave-inspired complex vectors. In Token2Wave, each token is represented with a magnitude component, capturing the global semantics of the entire input text, and a phase component, encoding the relationships between individual tokens and the global semantics. Building on prior research that demonstrated the effectiveness of wave-like operations, such as interference and modulation, during forward propagation, this study investigates the convergence behavior, backpropagation characteristics, and embedding independence within the Token2Wave framework. A detailed computational complexity analysis shows that Token2Wave can significantly reduce video memory usage and training time compared to BERT. Gradient comparisons for the [CLS] token, total input text, and classifier parameters further highlight Token2Wave's unique characteristics. This research offers new insights into wave-based token representations, demonstrating their potential to enable efficient and computationally friendly language model architectures.}

\maketitle

	\section{Introduction}\label{sec1}
	Currently, there are two types of token embedding methods. The fixed token embedding, such as Skip-gram and Continuous Bag of Words (CBOW) \cite{bib1}, assign the same embedding vector to each token, which cannot adapt to the dynamic meanings of tokens in varying contexts. The context-dependent embedding, on the other hand, generates different embeddings for the same token depending on its contexts. Many current Natural Language Processing (NLP) methods, such as the Transformer \cite{bib5}, use the attention mechanism to update token embeddings by measuring relationships between tokens with dot products. However, attention only infers global semantics indirectly through pairwise relationships rather than directly capturing the overall meaning of the text.
	
	In our previous work \cite{zhang2024wavenetworkultrasmalllanguage}, we introduced the Wave Network, a language model based on a new token representation method called Token2Wave. Token2Wave uses \textbf{complex vector token representations} to represent both the global and local semantics of each token with two parts: a magnitude vector representing the global semantics of the input text, and a phase vector capturing the relationships between individual tokens and global semantics. The \textbf{complex vector token representations} enables wave-like operations, such as interference and modulation for efficient updates.
	
	While the previous work focused on constructing token representations as waves and their forward propagation in text classification tasks, the current study delves into the architectural and functional aspects of the Wave Network. Here, we present a thorough analysis of the convergence performance, gradient behaviors of the network components (e.g., [CLS] embedding, overall input embedding, classifier), and the independence level among embedding dimensions. By focusing on these aspects, we aim to provide deeper insights into the theoretical details of the Wave Network and its potential effectiveness in various NLP tasks.
	
	\section{Representing Tokens as Waves}\label{sec2}
	
	In this framework, we represent each token using a complex vector in the form of $\mathbf{G} \cdot e^{i \cdot \boldsymbol{\alpha}}$, which comprises two components: a magnitude vector $\mathbf{G}$ that represent the global semantics of the text, and a phase vector $\boldsymbol{\alpha}$ that encode the relationships between individual tokens and the global context. We refer to this form of token representation as the complex vector token representation. 
	
	\noindent\textbf{1)} \textbf {Global Semantics Representation}\label{sec:global_semantics1}
	
	The meaning of each token within a text frequently relies on the overall meaning of the entire context. Representing these global semantics can help disambiguate individual tokens, making this understanding critical for downstream tasks that depend on a global view of the text.
	
	Based on principles from signal processing, where signals are often represented in polar coordinates, we treat each token as a discrete signal in the frequency domain. Here, magnitude represents the signal’s intensity, and phase specifies its relative position within a cycle \cite{Feynman_1494701}. As shown in the part (I) of Figure \ref{angle}, given an input text with $n$ tokens $input\_ text = [\mathbf{w}_1, \mathbf{w}_2, \dots,\mathbf{w}_j, \dots,\mathbf{w}_{n}]$:

	\begin{figure}[H]
		\vspace{-20pt}
		\centering
		\includegraphics[width=1.1\textwidth]{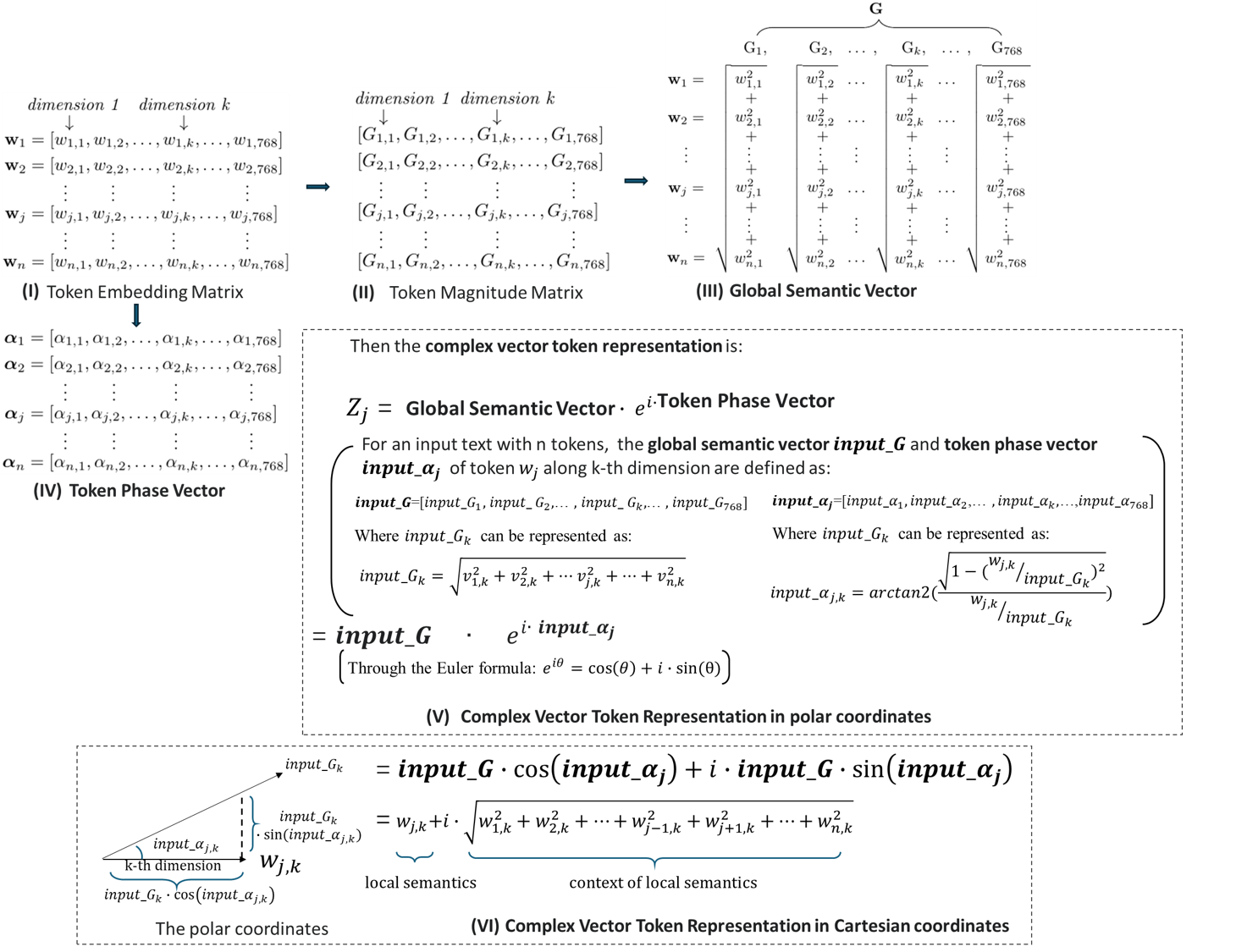}
		\caption{Create \textbf{complex vector token representations} from token embeddings}
		\label{angle}
		\vspace{-18pt}
	\end{figure}
	
	Each token embedding \( \mathbf w_{j} \) can be treated as a discrete real-value signal, where each elements $w_{j,k}$ represents the signal component along the $k$-th dimension. From a physical perspective, the magnitude of each signal component is defined as $G_{j,k}=|{w}_{j,k}|$, and the energy of each signal component can be defined as $E = {w}_{j,k}^2$ \cite{10.5555/248702}. Using these magnitudes, we construct the token magnitude matrix from the token embedding matrix, as illustrated in part (II) of Figure \ref{angle} and part (I) of Figure \ref{angle}. Next, as shown in the part (III) of Figure \ref{angle}, we sum the magnitudes of all token embedding components along each dimension to define the \textbf{global semantics vector} $\mathbf{G} = [{G}_1, {G}_2, \dots, {G}_k, \dots, {G}_{768}]$, where each \textbf{global semantic element} ${G_k}$ can be defined as $\text{G}_{k} = \left\| \mathbf{w}_{:,k} \right\|_2=\left\| [w_{1,k}, w_{2,k}, \dots,w_{j,k} , \dots,w_{n,k}] \right\|_2= \sqrt{{w_{1,k}^2 + w_{2,k}^2 + \dots +w_{j,k}^2 +  \dots+ w_{n,k}^2}}$. Here, \( w_{j,k} \) represents the \( k \)-th dimension of the \( j \)-th token embedding. This \textbf{global semantic vector} \( \mathbf{G} \) represents the global semantics of the entire input text and will serve as the magnitude of the \textbf{complex vector token representation} of each token in polar coordinates, as shown in the part (V) of Figure \ref{angle}. For simplicity, we focus on input-level global semantics in this research. Then, given a input text with $n$ token embeddings, the input-level global semantics vector can be defined as $\mathbf{input\_G} = \left[{input\_G}_1,{input\_G}_2, \dots,{input\_G}_k, \dots, {input\_G}_{768} \right]$, where ${input\_G}_{k} = \sqrt{{w_{1,k}^2 + w_{2,k}^2 +  \dots +w_{j,k}^2 +\dots+ w_{n,k}^2}}$.
	
	\noindent\textbf{2)} \textbf {Local Semantics Representation}\label{sec:global_semantics}
		
	Local semantics typically capture the specific meaning of each individual token, helping in the analysis of dependencies and fine-grained distinctions between tokens within a text. For instance, tasks such as sentiment analysis, entity recognition, and keyword extraction often rely on an accurate understanding of each token’s unique meaning.
	
	From a signal processing perspective, \textit{phase describes the relative relationships between signals. We will use the phase of \textbf{complex vector token representations} to represent the relative relationships between individual tokens and the \textbf{global semantic vector}.} That is, the phase representation of a token is coupled with its \textbf{global semantic vector}. For each token $\boldsymbol{w_{j}}$ in the input text, its \textbf{phase vector} is $\boldsymbol{\alpha_j} = \left[{\alpha_1}, {\alpha_2}, \dots,{\alpha_k}, \dots, {\alpha_{768}} \right]=\left[{input\_\alpha_1}, {input\_\alpha_2}, \dots,{input\_\alpha_k}, \dots, {input\_\alpha_{768}} \right]$, where $\mathit{input\_\alpha_k}$ is defined as $arctan2(\frac{\sqrt{1-(\frac{w_{j,k}}{{input\_G}_{k}})^2}}{\frac{w_{j,k}}{{input\_G}_{k}}})$ based on the corresponding element $\mathit{input\_G_k}$ in the \textbf{global semantic vector} of the input text $\mathbf{input\_G} = \left[{input\_G}_1,{input\_G}_2, \dots,{input\_G}_k, \dots, {input\_G}_{768} \right]$. Note that we use the function arctan2 to ensure angles fall within the range of $-\pi$ to $\pi$, consistent with the standard phase angle in physics. Using these definitions, we can derive the token phase matrix from the token embedding matrix, transform from part (I) to part (IV) in Figure \ref{angle}.
	
	To represent the \textbf{complex vector token representations} in Cartesian coordinates, we use the Euler's formula $e^{i\theta} = \cos(\theta) + i \cdot \sin(\theta)$ \cite{Ahlfors1966} to convert \textbf{complex vector token representations} from polar to Cartesian coordinates, as shown in part (VI) of Figure \ref{angle}. For example, the \textbf{complex vector token representations} $\mathbf{input\_G}$ can be expressed in Cartesian coordinates as $\mathbf{input\_G} \cdot \cos(\boldsymbol{input\_\alpha_j}) + i \cdot \mathbf{input\_G} \cdot\sin(\boldsymbol{input\_\alpha_j})$. The inner product of $\sin(\boldsymbol{input\_\alpha_j})$ and $\cos(\boldsymbol{input\_\alpha_j})$ is zero over a full period, making them orthogonal \cite{arfken2013mathematical}. Consequently, the real part $\mathbf{input\_G} \cdot \cos(\boldsymbol{input\_\alpha_j})$ and the imaginary part $i \cdot \mathbf{input\_G} \cdot \sin(\boldsymbol{input\_\alpha_j})$ are also orthogonal, fulfilling the properties of wave representations as described in physics \cite{jackson_classical_1999}. As Figure \ref{angle} illustrates, the real part of the token embedding $\boldsymbol{w_{j}}$ represents the token's contribution along the \( k \)-th dimension, capturing the local semantics of the input text. The imaginary part describes the \textbf{global semantic element} apart from $\boldsymbol{w_{j}}$, representing the context of token $\boldsymbol{w_{j}}$ along the $k$-th dimension within the input text. 

	\section{Complex Vector Token Representation Update}\label{sec:superposition}
	Complex vectors naturally align with the physical properties of waves \cite{berman2018introductory, herman2015introduction}, enabling the use of wave-inspired operations for efficient updates to complex vector token representations. In our Wave network, we introduce a linear layer designed to generate two distinct versions of each token’s complex vector representation at the input-text level. These versions facilitate the application of wave-based operations, such as interference and modulation.
	
	\subsection{Wave Interference} \label{waveinterference}
	From physical perspective, \textit{wave interference is a phenomenon where two coherent waves combined by adding their intensities or displacements, considering their phase difference.} In the context of generating \textbf{complex vector token representations} from input-level global semantics, as discussed in the Section \ref{sec2}, we define two variant \textbf{complex vector token representations} for token \( \boldsymbol{w_{j}} \) as: $\mathbf{input\_Z_{j}} = \mathbf{input\_G} \cdot e^{i \cdot \boldsymbol{input\_\alpha_j}}$ and $  \mathbf{input\_Z_{j}'} = \mathbf{input\_G'} \cdot e^{i \cdot \boldsymbol{input\_\alpha_j'}}$. We use complex vectors addition to simulate wave interference \cite{openstax2024university} and obtain the combined \textbf{complex vector token representation} $\mathbf{interference\_Z_{j}}$ for token $\boldsymbol{w_{j}}$ as follows:
	\setlength{\abovedisplayskip}{3pt}  
	\setlength{\belowdisplayskip}{3pt}  
	\begin{equation}
		\begin{split}
			&\mathbf{Interference\_Z_{j}} = \mathbf{input\_Z_{j}} + \mathbf{input\_Z_{j}'}= \mathbf{input\_G} \cdot e^{i \cdot \boldsymbol{input\_\alpha_j}} + \mathbf{input\_G'} \cdot e^{i \cdot \boldsymbol{input\_\alpha_j'}}\\
			&= \left( \mathbf{input\_G} \cdot \cos(\boldsymbol{input\_\alpha_j}) + \mathbf{input\_G'} \cdot \cos(\boldsymbol{input\_\alpha_j'}) \right) \\
			&+ i \cdot \left( \mathbf{input\_G} \cdot \sin(\boldsymbol{input\_\alpha_j}) + \mathbf{input\_G'} \cdot \sin(\boldsymbol{input\_\alpha_j'}) \right)\\
			&= w_{j,k} + w'_{j,k}+ i \cdot (\sqrt{{w_{1,k}^2 + w_{2,k}^2 + \dots+w_{j-1,k}^2 +w_{j+1,k}^2 +\dots+ w_{n,k}^2}}\\
			&+\sqrt{{w_{1,k}'^2 + w_{2,k}'^2 + \dots+w_{j-1,k}'^2 +w_{j+1,k}'^2 +\dots+ w_{n,k}'^2}})
		\end{split}
		\label{formula_interference5}
	\end{equation}
	
	Next, we illustrate how the phase difference between two complex vector token representations, such as \( \mathbf{input\_Z_{j}} \) and \( \mathbf{input\_Z_{j}'} \), affects the overall intensity of the resulting complex vector through their interference term. As discussed in detail in our prior work \cite{zhang2024wavenetworkultrasmalllanguage}, the interference term \( \text{Re}(\mathbf{input\_Z_j} \cdot \overline{\mathbf{input\_Z_j'}}) \) can be derived from the square of the magnitude of \( {input\_Z_{j,k}} \). This interference term indicates how the phase difference between two complex vector representations determines constructive or destructive interference. Briefly, we express the interference term as follows:
	
	\begin{equation}
		\begin{split}
			2 \cdot \text{Re} (\mathbf{input\_Z_j} \cdot \overline{\mathbf{input\_Z_j'}}) &= 2 \cdot \text{Re} \left(\mathbf{input\_G} \cdot e^{i \cdot \boldsymbol{input\_\alpha_j}} \cdot \mathbf{input\_G'} \cdot e^{-i \cdot \boldsymbol{input\_\alpha_j'}}\right) \\
			&= 2 \cdot \mathbf{input\_G} \cdot \mathbf{input\_G'} \cdot \cos(\boldsymbol{input\_\alpha_j} - \boldsymbol{input\_\alpha_j'})
		\end{split}
		\label{interferenceitem2}
	\end{equation}
	Equation \ref{interferenceitem2} demonstrates that the cosine value of the phase difference directly determines the interference result.
	
	\subsection{Wave Modulation}\label{sebsection:wavemodulation}
	From a physical perspective, \textit{wave modulation involves varying one or more characteristics of a periodic waveform, known as the carrier signal, in response to a separate input signal that contains the information to be transmitted.} In signal processing, this concept is applied in two main forms. First, \textit{amplitude modulation} adjusts the amplitude of the carrier wave based on the input signal’s amplitude, encoding information in the wave’s strength \cite{georgi1993physics}. Second, \textit{phase modulation} varies the phase of the carrier wave in response to the input signal’s changes, encoding information through shifts in the wave’s position \cite{phasemod}. Both amplitude and phase modulation can be achieved by multiplying complex vectors representing waves \cite{CRECRAFT2002200, PURSLEY200223, james2003digital}.
	
	In the context of generating \textbf{complex vector token representations} from input-level global semantics, as discussed in the Section \ref{sec2}, we consider two variant \textbf{complex vector token representations} for token \( \boldsymbol{w_{j}} \) as: $\mathbf{input\_Z_{j}} = \mathbf{input\_G} \cdot e^{i \cdot \boldsymbol{input\_\alpha_j}}$ and $  \mathbf{input\_Z'_{j}} = \mathbf{input\_G'} \cdot e^{i \cdot \boldsymbol{input\_\alpha_j'}}$. We use complex vectors multiplication to simulate wave modulation \cite{openstax2024university} and obtain the combined \textbf{complex vector token representation} $\mathbf{modulation\_Z_{j}}$ for token $\boldsymbol{w_{j}}$ as follows. 
	
	\begin{equation}
		\begin{split}
			&\mathbf{Modulation\_Z_{j}} = \mathbf{input\_Z_{j}} \cdot \mathbf{input\_Z_{j}'}= \mathbf{input\_G} \cdot e^{i \cdot \boldsymbol{input\_\alpha_j}} \cdot  \mathbf{input\_G'} \cdot e^{i \cdot \boldsymbol{input\_\alpha_j'}}  \\
			&= \mathbf{input\_G} \cdot \mathbf{input\_G'} \cdot e^{i \cdot \boldsymbol{input\_\alpha_j} + \boldsymbol{input\_\alpha_j'}}\\
			&=\mathbf{input\_G} \cdot \mathbf{input\_G'} \cdot \cos(\boldsymbol{input\_\alpha_j} + \boldsymbol{input\_\alpha_j}')\\
			&+ i \cdot \mathbf{input\_G} \cdot \mathbf{input\_G'} \cdot \sin(\boldsymbol{input\_\alpha_j}+ \boldsymbol{input\_\alpha_j'})\\
			&= (w_{j,k}\cdot w_{j,k}'-\sqrt{{w_{1,k}^2 + w_{2,k}^2 + \dots+w_{j-1,k}^2  +w_{j+1,k}^2 + \dots +w_{n,k}^2}} \\
			&\cdot \sqrt{{w_{1,k}'^2 + w_{2,k}'^2 + \dots+w_{j-1,k}'^2 +w_{j+1,k}'^2 +\dots+ w_{n,k}'^2}})\\
			&+ i \cdot  \left(w_{j,k}' \cdot \sqrt{{w_{1,k}^2 + w_{2,k}^2 + \dots+w_{j-1,k}^2 + w_{j+1,k}^2 + \dots +w_{n,k}^2}}\right.\\
			&\left.+ (w_{j,k} \cdot \sqrt{{w_{1,k}'^2 + w_{2,k}'^2 + \dots+w_{j-1,k}'^2 +w_{j+1,k}'^2 +\dots + w_{n,k}'^2}}\right)
		\end{split}
		\label{formula_modulation555}
	\end{equation}
	
	For detailed computation steps, please refer to our previous work \cite{zhang2024wavenetworkultrasmalllanguage}.

	\subsection{Restore Token Embedding from Complex Vector Token Representations}
	BERT utilizes a special classification embedding [CLS] at the beginning of each input text to represent the overall input, and then make predictions based on this [CLS] embedding when performing text classification tasks \cite{yu2019improving}. To facilitate an accurate comparison with the Transformer and BERT on token representation and representation update, we also utilize a [CLS] token at the beginning of each input text to represent the overall input. the [CLS] token is then represented as a \textbf{complex vector token representation} like other tokens in the input text. For text classification tasks, we convert the representation of [CLS] embedding back to the token embedding space along with other tokens.
	
	As described in Section \ref{sec2}, we restore the token embeddings by performing a multiplication between the  \textbf{global semantic vector} $\mathbf{input\_G}$ and the cosine value of the \textbf{phase vector} $\boldsymbol{input\_\alpha_j}$.

	\section{Experiments Settings}

	To ensure consistency and enable direct comparison, the experimental settings in this paper follow the same parameters as those used in our previous study \cite{zhang2024wavenetworkultrasmalllanguage}. Specifically, the learning rate is set to 1e-3 for both the Wave network and Transformer, and 2e-5 for BERT \cite{sun2019fine}. The batch size varies depending on the task: for resource utilization comparison experiments, all models use a batch size of 64; for accuracy comparison experiments, the batch size is 64 for the Wave network and Transformer, and 32 for BERT \cite{sun2019fine}. In the gradient comparison and embedding independence experiments, all three models use a batch size of 32. All models are trained or fine-tuned for four epochs. For fast convergence experiments, we evaluate test accuracy every 10 batches over a total of 500 batches. To maintain a consistent architecture, both the Wave network and Transformer use a single-layer structure. The Wave network generates initial token embeddings randomly using torch.nn.embedding in Pytorch \cite{paszke2019pytorch}, whereas the Transformer uses pre-trained BERT token embeddings. The BERT model is fine-tuned using the pre-trained base version.
	
	In the gradient and independence analysis experiments, we compare the Wave network and Transformer by focusing on three key components: (1) the back-propagation gradient of the cross-entropy loss of the [CLS] token embedding, (2) the back-propagation gradient of the cross-entropy loss of the overall token embeddings in the input text, and (3) the back-propagation gradient of the cross-entropy loss for the classifier parameters. By excluding auxiliary components like linear and feed-forward layers, we ensure a controlled comparison of how core components respond to changes in training data and influence learning dynamics. Additionally, we analyze the feature independence across the dimensions of the [CLS] token embedding to better understand model behavior.
	
	We collected data for this analysis by training on the AG News dataset, which contains 96,000 training samples. With a batch size of 32, each epoch consists of 3,000 batches, and all other parameters remain consistent with previous experiments. To capture trends across all epochs, we plotted performance data with the x-axis representing batches and distinct colored lines showing performance in each epoch. Additionally, we used Kernel Density Estimation (KDE) \cite{davis2011remarks} \cite{parzen1962estimation} to gain deeper insights into dynamic training behavior and gradient optimization paths.
	
	In this context, the gradient norm refers to the magnitude of the gradient, which reflects the extent of parameter updates during training. It indicates how much the model adjusts its parameters during backpropagation. Thus, when discussing changes in the gradient, we are actually describing how the model updates weights in each training step to minimize cross-entropy loss. Therefore, in this section, the gradient norm is consistently used to represent the amplitude of parameter or weight adjustments.
	
	\section{Result and Analysis}
	\subsection{Accuracy Comparison}
	\small
	\begin{table*}[t]
		\centering
		\begin{minipage}{0.48\textwidth}
			\centering
			\begin{minipage}{\textwidth}
				\centering
				\begin{tabular}{>{\centering\arraybackslash}m{1cm}
						>{\centering\arraybackslash}m{1.5cm}
						>{\centering\arraybackslash}m{1.1cm}
						>{\centering\arraybackslash}m{1.5cm}
						>{\centering\arraybackslash}m{1cm}}
					\toprule
					Model & Data sets & Acc. & VRAM(GB) & Time(s)  \\
					\midrule
					WNI & AG News & 90.36\% & 0.30 & 146.90\\
					WNM & AG News & 91.29\% & 0.32 & 147.02\\
					TF & AG News & 71.68\% & 0.85 & 173.25 \\
					\bottomrule
				\end{tabular}
				\label{tab:performance_comparison}
			\end{minipage}
			\vfill
			\begin{minipage}{\textwidth}
				\centering
				\includegraphics[width=1.1\linewidth]{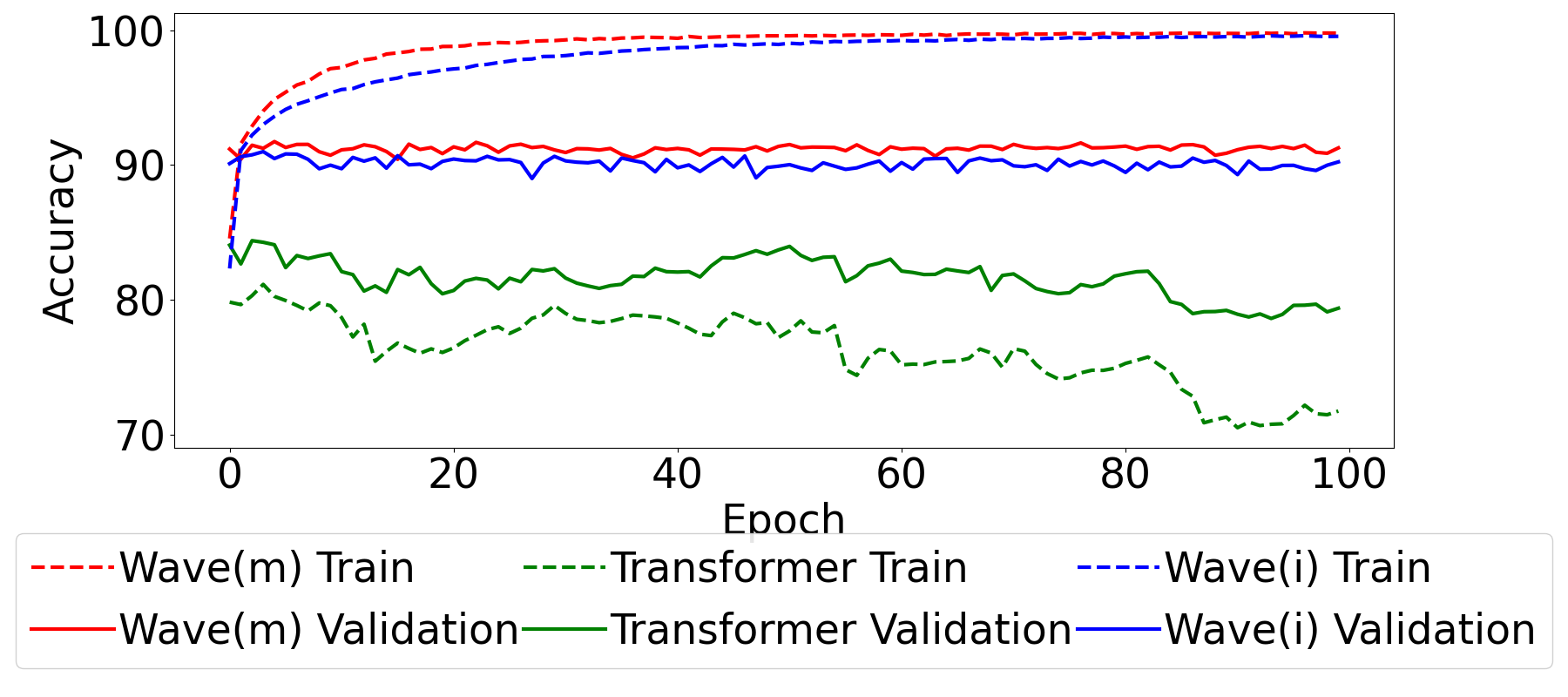}
				\caption{Compare the Wave network and Transformer}
				\label{fig:additional_figure}
			\end{minipage}
		\end{minipage}
		\hfill
		\begin{minipage}{0.48\textwidth}
			\centering
			\begin{tabular}{>{\centering\arraybackslash}m{1cm}
					>{\centering\arraybackslash}m{1.5cm}
					>{\centering\arraybackslash}m{1.1cm}
					>{\centering\arraybackslash}m{1.5cm}
					>{\centering\arraybackslash}m{1cm}}
				\toprule
				Model & Data sets & Acc. & VRAM(GB) & Time(s) \\
				\midrule
				WNI & AG News & 90.91\% & 0.30 & 146.90 \\
				WNM & AG News & 91.66\% &  0.32 & 147.02 \\
				BERT & AG News & 94.64\% & 1.28 & 1034.99\\
				WNI & DBpedia14 & 97.93\% & 0.30 & 979.78 \\
				WNM& DBpedia14 & 98.05\% & 0.30 & 991.15 \\
				BERT & DBpedia14 & 99.28\% & 1.27 & 2734.76  \\
				WNI & IMDB & 87.00\% & 0.37 & 119.57\\
				WNM & IMDB & 87.02\% & 0.37 & 119.96 \\
				BERT & IMDB & 93.94\% & 1.27 & 220.46 \\
				\bottomrule
			\end{tabular}
			\caption{Compare the Wave network and BERT base}
			\footnotesize
			\textbf{Note:} {WNI: Wave network with Interference; WNM: Wave network with Modulation; TF: Transformer}\label{WNIWNM}
			\label{tab:performance_comparison_2}
		\end{minipage}
		\vspace{0.5cm} 
		\vspace{-30pt}
	\end{table*}
	\normalsize
	
	This section replicates the accuracy comparison from our previous work \cite{zhang2024wavenetworkultrasmalllanguage}, verifying similar improvements in VRAM usage, training time, and convergence speed. As shown in Table \ref{fig:additional_figure} and Table \ref{tab:performance_comparison_2} for the AG News dataset, compared to the single-layer Transformer, the WNI and WNM in Table \ref{WNIWNM} significantly reduces the VRAM consumption by 64.71\% and 62.35\%, respectively, and shorten training time by 15.21\% and 15.14\%. At the same time, they improve classification accuracy by 18.68\% and 19.61\%, corresponding to wave interference and modulation. When compared to the pre-training BERT base, the single-layer Wave network reduces VRAM consumption by 76.56\% and 75\%, and cuts training time by 85.8\%, while maintaining 96.96\% and 96.85\% of BERT's accuracy.
	
	\subsection{Convergence Performance}
	
	We hypothesize that effective representation and update methods should enable models to complete downstream tasks with minimal training, even when starting from randomly initialized embeddings. To test this, we collected 50 data points from the Wave network and Transformer, assessing test set accuracy after every ten training batches.

	\begin{figure*}[ht]
		\centering
		\begin{minipage}{0.4\textwidth}
			\centering
			\begin{tabular}{>{\centering\arraybackslash}m{0.7cm}
					>{\centering\arraybackslash}m{0.9cm}
					>{\centering\arraybackslash}m{0.7cm}
					>{\centering\arraybackslash}m{0.9cm}}
				\toprule
				Batch & Acc. & Batch & Acc. \\
				\midrule
				10 & 25.00\% & 110 & 50.01\% \\
				20 & 25.00\% & 120 & 55.84\%  \\
				30 & 25.00\% & 130 & 67.01\%  \\
				40 & 25.00\% & 140 & 68.75\%  \\
				50 & 25.00\% & 150 & 76.39\%  \\
				60 & 25.00\% & 160 & 74.38\%  \\
				70 & 25.00\% & 170 & 75.87\%  \\
				80 & 30.55\% & 180 & 81.64\%  \\
				90 & 33.97\% & 190 & 81.99\%  \\
				100 & 37.57\% & 200 & 84.25\%  \\
				\bottomrule
			\end{tabular}
			\label{tab:convergence}
		\end{minipage}%
		\hfill
		\begin{minipage}{0.6\textwidth}
			\centering
			\includegraphics[width=\linewidth]{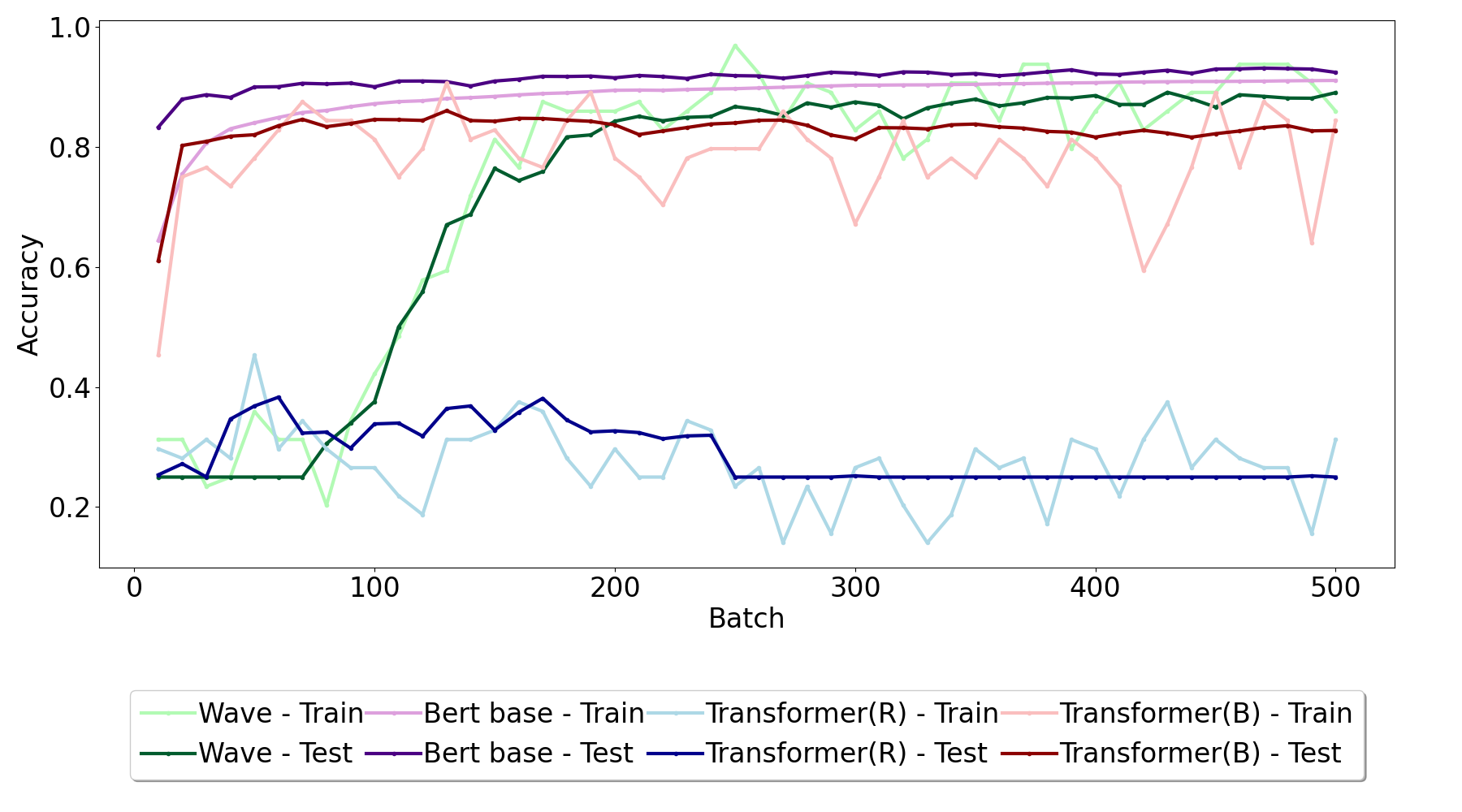}
			\label{fig:quickerconvergence}
			
		\end{minipage}
		\textbf{Note:} Transformer(R): Transformer with randomly token embeddings; Transformer(B): Transformer with BERT token embeddings
		\caption{Quick Convergence on AG News}
		\label{fig:combined_figure}
		
	\end{figure*}
	
	Table \ref{fig:combined_figure} presents accuracy data for the single-layer Wave network after every ten training batches during the first 200 batches. Figure 7 compares the convergence performance of the Wave network, Transformer, and BERT base. Both results clearly indicate that the Wave network begins to converge around the 100th batch, achieving high accuracy by the 200th batch. This demonstrates that after processing 12,800 samples, the model's dimensions approximate correct global semantics, even when starting from randomly initiated embeddings.
	
	These findings show that holistic semantics-based word embeddings can effectively learn semantic representations that are sufficient for classification tasks, even with minimal training.

	\subsection{Gradients of [CLS] Embedding:} \label{clsback}
		\begin{figure*}[ht]
	\centering
	\includegraphics[width=0.9\textwidth, height=0.9\textheight, keepaspectratio]{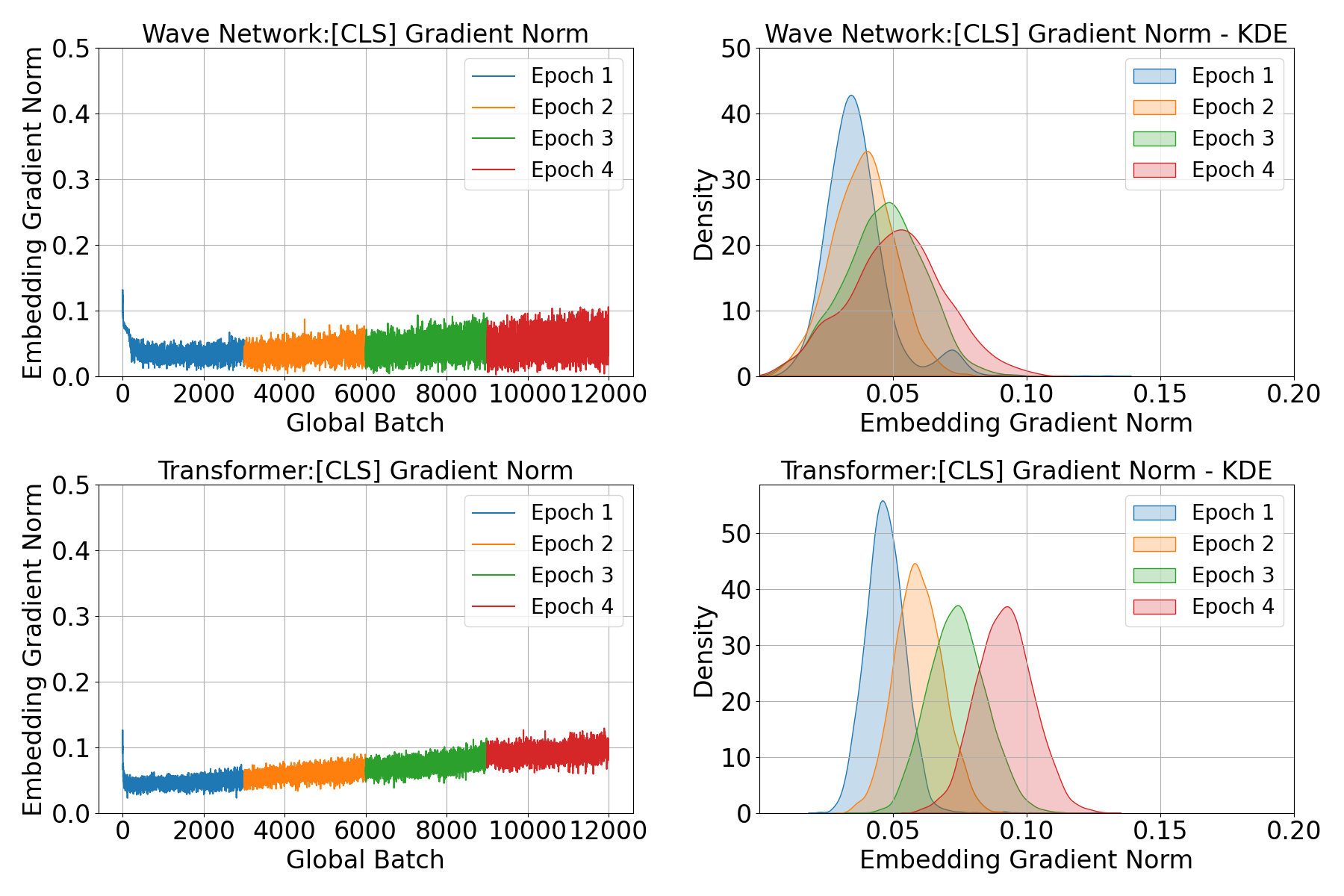}
	\caption{Compare the gradients of [CLS] embedding between the Wave network and Transformer}
	\label{fig:gradient}
	\end{figure*}
	For the Wave network's [CLS] token embedding gradient norm graph, the [CLS] token summarizes information from the entire input text. At the start of the first epoch, the gradient norm dropped sharply from 0.13 to an oscillation range between 0.05 and 0.02. This sharp drop indicates that the [CLS] token embedding initially struggled to capture the necessary information for classification, prompting significant adjustments. In subsequent training, the minimum gradient value stabilized between 0.01 and 0.02, while the maximum gradually increased from 0.05 to 0.1. This suggests that while the of the [CLS] token embedding of most samples were adjusted early, substantial adjustments continued for certain difficult-to-classify samples during the later stages of training.
	
	For the Transformer's [CLS] token embedding gradient norm graph, there was a similarly large adjustment during early training, akin to the Wave network. Over four epochs, the gradient norm showed a consistent upward trend. During the last three epochs, both the minimum and maximum gradient values steadily increased, with the minimum rising from 0.03 to 0.085 and the maximum from 0.05 to 0.115. This indicates that, while the model gradually stabilized, the Transformer continued fine-tuning the [CLS] token representation, particularly for certain difficult-to-classify samples. Compared to the Wave network, the Transformer's gradient changes were more stable and exhibited a narrower range, reflecting its focus on more local semantic adjustments caused by the dot product between token embeddings, without being affected by the entire token embeddings in the input text like \textbf{complex vector token representations}.
	
	For the [CLS] token embedding's norm KDE graph of the Wave network, the first epoch shows significant long-tail characteristics, meaning that the [CLS] gradient norm of some samples extended into a larger range. This indicates that in the early stages of training, the model had large prediction errors for a small subset of samples, requiring substantial gradient adjustments to reduce their cross-entropy loss. At the same time, the gradient norm of most samples concentrated at smaller values (e.g., the peak in the first epoch is 0.03), showing that the Wave network quickly reduced prediction errors for the majority of samples, efficiently lowering their cross-entropy loss early in the training. From the second to the fourth epoch, the peak of the [CLS] gradient norm shifts slightly to the right (from 0.03 to 0.055), indicating a slight increase in gradient norm values. However, the peaks of the last three epochs all lie on the descending slope of the previous epoch, suggesting that while the gradient norm increased for harder samples, the overall weight updates became progressively smaller. This shows that the Wave network experienced fewer prediction errors as training progressed, leading to smaller overall gradient updates. The classification task stabilized, and gradient updates were increasingly concentrated on a few hard-to-classify samples. Overall, the gradient distribution of the last three epochs became more compact, and aside from the long-tail behavior in the first epoch, the norm range remained consistent. This suggests that the Wave network adapted to most samples through global adjustments early in training, with later epochs focusing on fine-tuning to improve the accuracy of more difficult samples.
	
	For the [CLS] token embedding's norm KDE graph of the Transformer, the peaks of the [CLS] gradient norm steadily shift to the right from 0.04 in the first epoch to 0.09 in the fourth epoch, with each peak higher than those of the Wave network across all epochs. The peaks are also more distinctly separated, indicating clearer differences in gradient norms between epochs. Interestingly, the peaks in the last two epochs are nearly identical, suggesting that the model stabilizes in later stages, but continues to require significant gradient updates for certain samples. While the expansion in the gradient norm range is slower compared to the Wave network, the Transformer shows a more pronounced rightward shift in the peaks, leading to a higher concentration of large gradient norms in later epochs. Across all four epochs, the distribution of the Wave network remains broader, particularly in the first epoch where its right-tail norm extends beyond that of the Transformer. In the second epoch, the two models show similar maximum norm values. By the third and fourth epochs, the Transformer’s gradient norm distribution shifts further to the right, with a larger proportion of higher gradient norms compared to the Wave network, indicating that the Transformer maintains larger gradient updates, especially for harder-to-classify samples. Despite the Wave network having a consistently broader distribution, the Transformer exhibits a more significant increase in gradient norms in later training stages, reflecting its continuous adjustments. This pattern suggests that while the Wave network undergoes wider global adjustments early in training, the Transformer progressively concentrates on larger norm values in later epochs, particularly in the third and fourth epochs. This leads to a more focused but intense gradient adjustment process, contrasting with the broader but more gradual adjustments seen in the Wave network.
		\begin{figure}[h]
		\centering
		\includegraphics[width=0.7\linewidth]{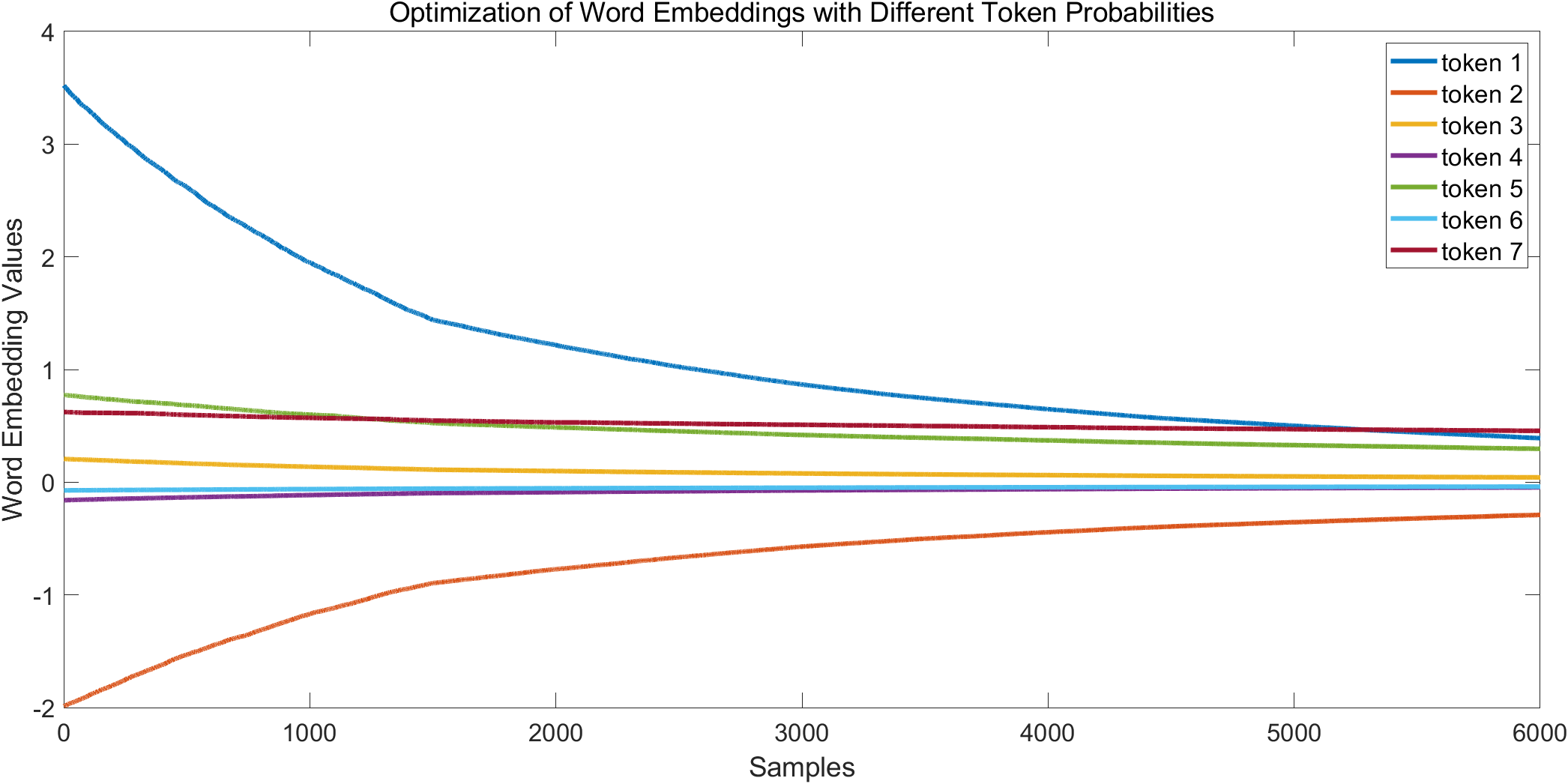}
		\caption{Construct wave representation by complex number}
		\label{quantityofback}
	\end{figure}
	
	When comparing the handling of difficult-to-classify samples, the Wave network exhibits significant long-tail characteristics in the first epoch, suggesting that a subset of challenging samples undergoes substantial gradient updates to reduce cross-entropy errors. This behavior can be attributed to the Wave network's use of magnitude and phase representations in polar coordinates, which quickly capture global features and enable rapid adjustments for difficult samples. In the last three epochs, while the peak values decrease, the right tail of the norm distribution remains stable or slightly increases, indicating that the gradient updates are distributed over a wider range. This suggests that although the majority of samples have been effectively handled, the model continues to make substantial adjustments for certain difficult-to-classify samples, focusing on refining their representations through fine-tuning. In contrast, the Transformer exhibits a wider gradient update range, and the magnitude of updates gradually increases throughout the training process. This suggests that the Transformer consistently addresses difficult-to-classify samples through incremental and uniform adjustments, progressively optimizing the [CLS] token embedding while maintaining stable gradient updates across most samples. From the perspective of convergence speed and stability, the Wave network converges faster due to its large early-stage gradient adjustments on difficult samples, which rapidly reduce cross-entropy errors for the majority of samples. On the other hand, the Transformer converges more slowly, with gradual and smooth gradient updates. This stable adjustment process allows the Transformer to effectively optimize cross-entropy errors throughout the entire training process. Consequently, the Transformer may outperform the Wave network when dealing with complex long texts or particularly challenging samples.
	
	It is worth mentioning that we conducted a simulation to illustrate how backpropagation affects token embeddings by isolating irrelevant layers (e.g., feedforward and linear layers) and focusing solely on the effect of wave modulation on token embedding backpropagation. In PyTorch's default backpropagation, the gradient descent process is expressed as $new\_w_{j,k} = old\_w_{j,k} - \eta \cdot \nabla w_{j,k} \cdot \textit{error},$ where \( new\_w_{j,k} \) is the updated token embedding, \( old\_w_{j,k} \) is the current token embedding, \( \eta \) is the learning rate, \( \nabla w_{j,k} \) is the gradient from wave modulation as defined in \ref{sub:backpropagation}, and \( \textit{error} \) represents the cross-entropy loss. For text classification tasks, CrossEntropy Loss is used to measure the difference between the model’s predicted and target distributions. Since PyTorch supports only real-value backpropagation, the gradient simplifies to \( 2w_{j,k} \), resulting in the update equation $new\_w_{j,k} = old\_w_{j,k} - \eta \cdot 2 \cdot old\_w_{j,k} \cdot \textit{error},$ which simplifies further to $new\_w_{j,k} = old\_w_{j,k}(1 - 2\eta \cdot \textit{error}).$ This leads to an exponential decay $new\_w_{j,k} = old\_w_{j,k}(1 - 2\eta \cdot \textit{error})^{iterations}.$ As the number of iterations increases, \( new\_w_{j,k} \) gradually decreases, as shown in Figure \ref{quantityofback}. We used the same experimental settings as in previous experiments, with 4 epochs and 1500 batches per epoch, totaling 6000 batches. The initial word embeddings follow \( \mathcal{N}(0, 1) \), meaning 68\% of data falls within one standard deviation from the mean, 95\% within two, and 99.7\% within three standard deviations\cite{degroot2013probability}. We simulated seven tokens with different average occurrence frequencies, ranging from 70\% for token 1 to 10\% for token 7. This implies that during backpropagation, tokens are updated at different iterations within the same batch. Using the average loss values from the first four epochs of actual training as the \textit{error} (0.4311, 0.2502, 0.2065, 0.1722), and applying the backpropagation formula, we approximated the gradient norm magnitude for the overall token embedding. The average length of samples in the AG News dataset was 37.85 tokens. Given an initial token embedding value of 1.0, applying the formula $new\_w_{j,k} = old\_w_{j,k}(1 - 2\eta \cdot \textit{error})^{iterations},$ with a learning rate of \( 1e-3 \) and an average loss of 0.4311 in the first epoch, we calculated the overall token embedding for a sample after backpropagation to be approximately 0.0326. Both the Wave network and Transformer use the [CLS] token for classification, meaning that during backpropagation, the cross-entropy loss is first propagated to the [CLS] token in the modulation or attention layer. In the Wave network, according to the chain rule of backpropagation, the cross-entropy loss passed to the [CLS] token further propagates through Formula \ref{formula_modulation5} in Subsection \ref{sub:backpropagation}, updating the overall token embedding. Based on the \textbf{complex vector token representation} shown in Figure \ref{angle}, we conclude that the backpropagation effect on the [CLS] token embedding is of the same order of magnitude as the overall token embedding. Additionally, this can explain the magnitude of the gradient norm in the wave network as shown in Figure \ref{fig:gradient} in this section and Figure \ref{fig:gradient1} in the next section.
	
	\subsection{Gradients of Overall Embedding of Input Text:}
	For the Wave network's overall token embedding gradient norm of the input text, at the beginning of the first epoch, the gradient norm oscillates between 0.01 and 0.04. This suggests that, in the early stages of training, the model makes global adjustments to reduce classification loss. Since classification errors vary greatly across samples, the overall gradient fluctuates accordingly. As training progresses, the maximum gradient norm increases from 0.04 to 0.09, indicating that some harder-to-classify samples have larger cross-entropy errors. As a result, the model applies larger gradients to adjust their token embeddings. Meanwhile, the minimum gradient norm remains at 0.01, showing that correctly classified samples consistently maintain small gradients.
	\begin{figure*}[ht]
		\centering
		\includegraphics[width=0.9\textwidth, height=0.9\textheight, keepaspectratio]{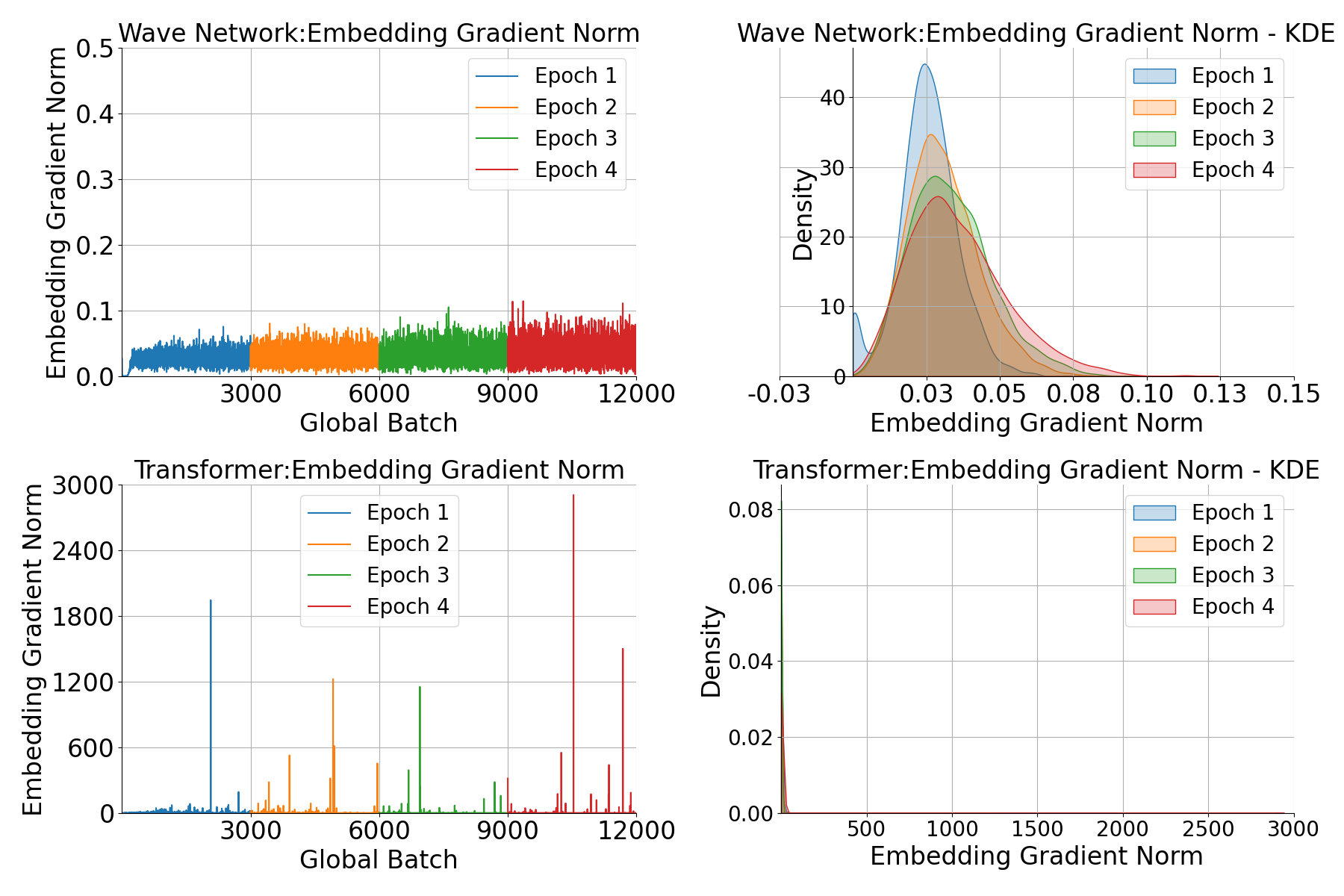}
		\caption{Compare the overall L2 norm of gradient tensor of word embeddings between wave network and the Transformer}
		\label{fig:gradient1}
	\end{figure*}
	For the Transformer's overall token embedding gradient norm, at around 2000 batches in the first epoch, the gradient norm peaks at 2000. Similar high peaks are observed in subsequent epochs, with values of 1200 in the second and 2900 in the fourth epoch. These extreme values suggest that certain samples have abnormally large cross-entropy errors, leading to very large gradients. Additionally, the presence of lower gradient norms for other samples indicates that errors persist across a wide range of samples, resulting in fluctuating token embeddings and inconsistent classification performance.

	In the Wave network's gradient norm KDE graph, the peak values across the four epochs are tightly clustered between 0.029 and 0.031, indicating that for most samples, classification errors change minimally. The model maintains stability in gradient updates for the majority of samples. However, the tail of the distribution extends from 0.07 in the first epoch to 0.12 in the fourth, showing that a small number of difficult samples still generate large errors, requiring more substantial updates.

	For the Transformer's gradient norm KDE graph, the peak value across the four epochs is around 0.075, indicating that the gradient norm for most samples is concentrated within a narrow range. However, by the fourth epoch, the tail of the distribution extends up to 3000, suggesting that a few samples still have abnormally large errors, despite ongoing training.
	\begin{figure*}[ht]
		\centering
		\includegraphics[width=0.9\textwidth, height=0.9\textheight, keepaspectratio]{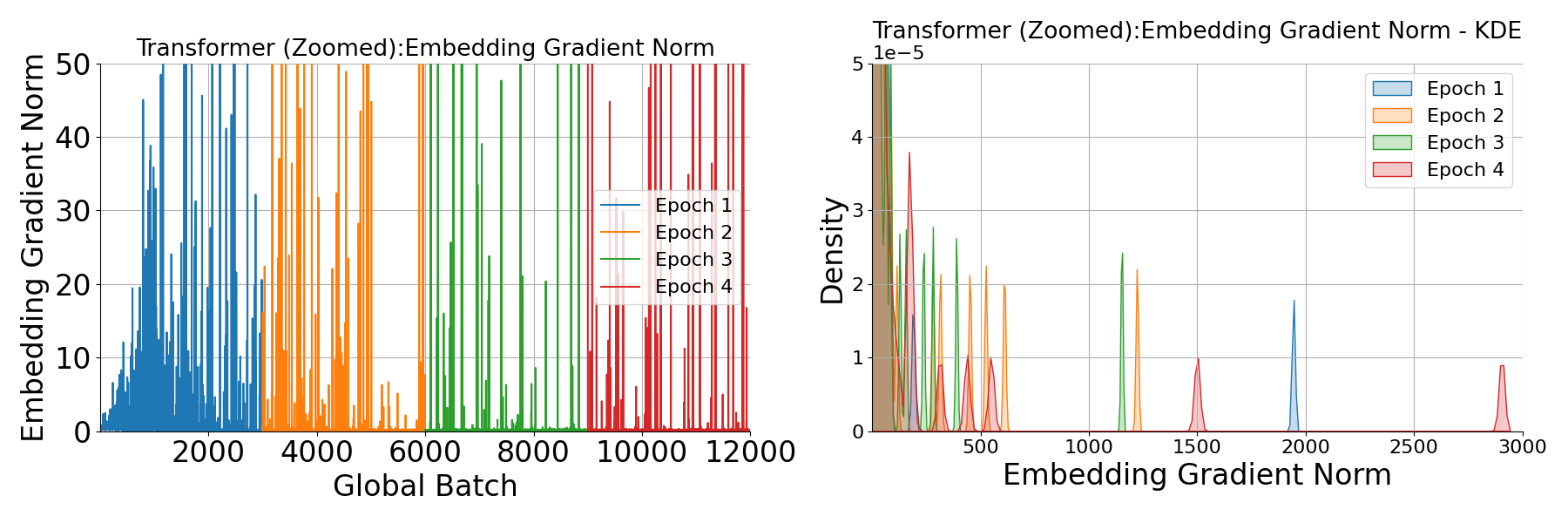}
		\caption{Zoomed overall L2 norm of gradient tensor on the Transformer's word embeddings}
		\label{fig:gradient2}
	\end{figure*}
	
	A comparison of the overall token embedding gradient norms reveals fundamental differences between the two embedding and updating methods. In the Wave network, as shown in Figure \ref{fig:gradient1}, the gradient of the [CLS] token embedding is propagated to the overall token embeddings through magnitude and phase in polar coordinates. The magnitude represents the global semantics of the input text, while the phase describes the angle between each token and the global semantics. Since the Wave network emphasizes global information transmission, during backpropagation, the [CLS] token's gradient is distributed across all token embeddings based on their contribution to the global semantics. This results in smaller differences in gradient updates between tokens, as their contributions are more evenly distributed, leading to a narrower gradient range and a shorter tail in the KDE graph. In contrast, the Transformer uses softmax weights to calculate attention between tokens. During backpropagation, the [CLS] token's gradient is distributed to other tokens based on their attention weights. Tokens with higher attention weights receive larger gradient updates, while those with lower weights receive fewer updates, as shown in Figure \ref{fig:gradient2}. This results in greater variation in the gradient distribution, reflected by a wider gradient range and a longer tail in the KDE graph. Additionally, the Transformer's focus on local semantics, especially for difficult samples, means certain tokens receive disproportionally large gradients in later epochs, resulting in the long-tail effect seen in the KDE graph.
	
	\subsection{Gradients of Classifier:}
		
	For the gradient norm graph of the classifier parameters in the Wave network, the gradient norm initially drops sharply from 28 to a range of 14 to 5 during the beginning of the first epoch. This indicates that the model made substantial adjustments to the classifier weights in the early stage of training, reflecting the need to reduce the initially high classification cross-entropy loss. Since this is a single-layer classifier, such large adjustments directly respond to the classification error's impact on the weights. As training progresses, the gradient norm gradually decreases, stabilizing within an oscillation range of 2 to 5. This suggests that after the initial reduction in classification error, weight updates become smaller and more stable, showing that the model's error gradually decreases. By the end of training, the lowest gradient value approaches 0, and the highest value falls to around 1, showing that the classifier weights require minimal adjustment and the model is nearly converged. Although there are notable fluctuations during the early and middle training stages, the gradient norm consistently decreases towards the end, indicating that the Wave network requires fewer adjustments to the classifier weights compared to the Transformer, suggesting that the classification error is almost fully optimized.
	\begin{figure*}[ht]
		\centering
		\includegraphics[width=\textwidth, height=\textheight, keepaspectratio]{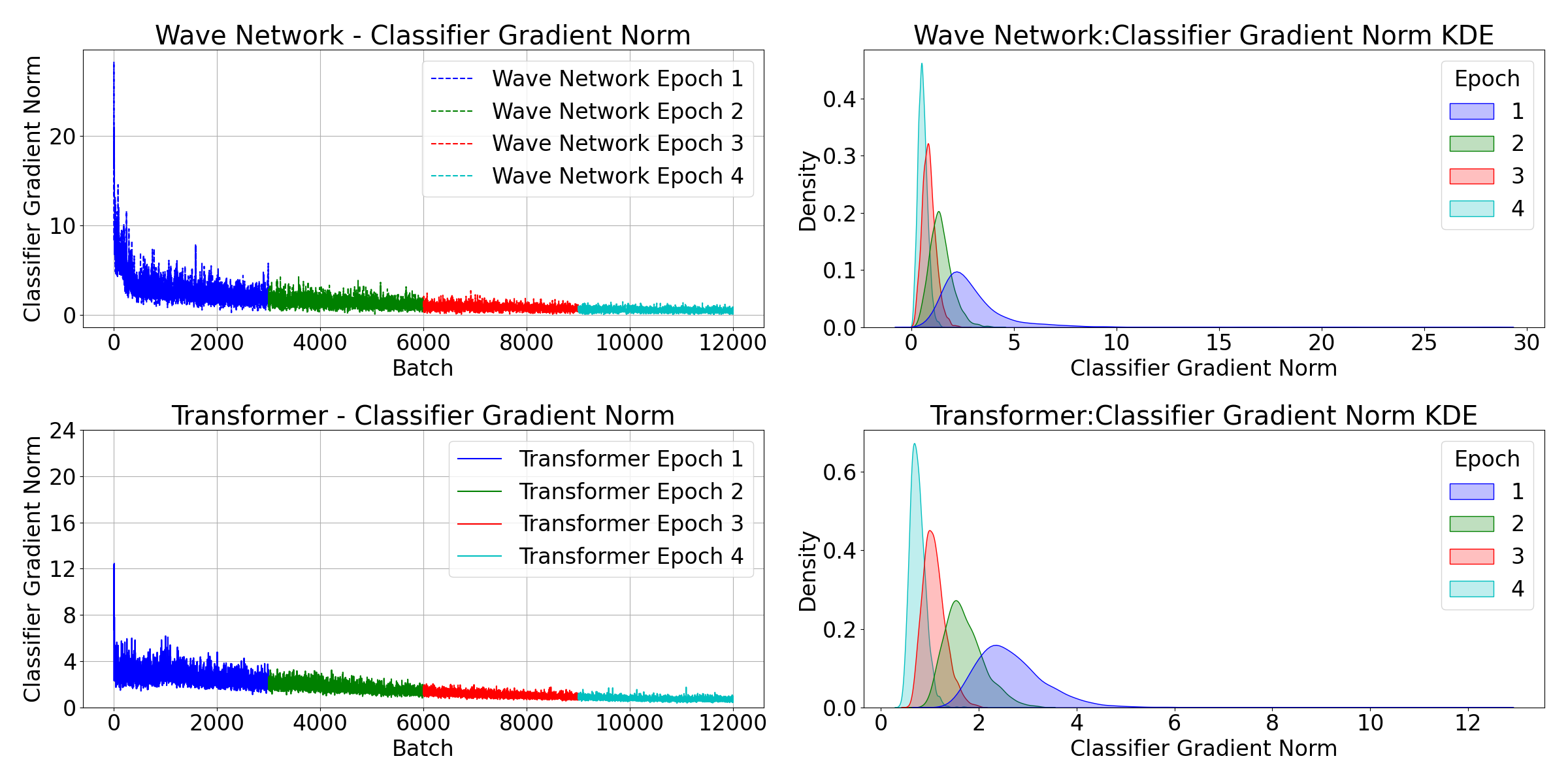}
		\caption{Compare the overall L2 norm of gradient tensor of classifier between wave network (modulation) and the Transformer}
		\label{fig:fullpage}
	\end{figure*}
	
	For the gradient norm graph of the classifier parameters in the Transformer, the gradient decreases from 13 to an oscillation range of 2 to 5.5 at the beginning of the first epoch. This reflects that the model makes smaller, more stable adjustments to the classifier weights early in training compared to the Wave network. The smoother initial adjustments suggest that the Transformer makes smaller updates when addressing classification errors. As training continues, the gradient norm further decreases, with the lowest value dropping to around 0.8 and the highest to 1.5, indicating a gradual reduction in cross-entropy loss and smaller adjustments to the classifier weights. The Transformer's self-attention mechanism allows it to evenly distribute relationships between input tokens, enabling smoother weight updates. However, this gradual adjustment means it takes longer to address errors from difficult-to-classify samples compared to the Wave network. As a result, while the Transformer's updates are smaller and more consistent, it takes more time to reach convergence similar to that of the Wave network.
	
	For the KDE of the gradient norm graph of the Wave network, the peak in the first epoch appears around 0.1, corresponding to a gradient norm about 2, indicating that most updates at the start of training are relatively moderate. However, the long tail extending to 29 reveals that the model applies significant gradient updates to a small number of difficult-to-classify samples at this early stage. As training progresses, the distribution of gradient updates shifts, and by the fourth epoch, the peak rises to 0.45. This indicates that gradient updates have become more concentrated for a larger portion of the samples, while the magnitude of updates decreases, showing some challenging samples still require optimization. The pronounced long tail in the first epoch suggests that the Wave network initially focuses on making large adjustments for a few samples with high classification errors, while most updates remain relatively small. As the model refines its learning in subsequent epochs, gradient updates become more concentrated, and the long tail shortens significantly. By the fourth epoch, most classification errors have been effectively addressed, with only a few difficult samples still requiring substantial updates, as indicated by the reduced length of the tail. This evolution of the gradient norm distribution across epochs reflects a dual optimization strategy employed by the Wave network. In the early stages, the model makes broad, significant updates to reduce errors across all samples, especially for hard-to-classify examples. As training progresses, updates become more focused and stable, with smaller adjustments concentrated on a smaller set of challenging samples. This shift allows the model to fine-tune its performance while maintaining overall stability in the later stages.
	
	For the KDE of the gradient norm graph of the Transformer, the peak in the first epoch appears at 0.18, corresponding to a gradient norm about 2.3, indicating that the majority of gradient updates early in training are relatively small. However, the long tail in the gradient distribution extends up to 13, suggesting that the Transformer applies significant updates to a small subset of difficult-to-classify samples. As training progresses, the peak shifts gradually to the left, reflecting a concentration of updates around smaller values, as the model becomes better at minimizing classification errors for most samples. By the fourth epoch, the peak increases to 0.68 while gradient norm decreases to around 1, indicating that the gradient updates, while more concentrated, still reflect larger adjustments as the model focuses on refining the remaining classification errors. The continued presence of a long tail, though less pronounced than in the first epoch, shows that some difficult-to-classify samples still require substantial updates. This pattern suggests that the Transformer makes large, targeted adjustments in the early stages of training to correct major classification errors, followed by more stable, focused updates as the model converges. However, even in later epochs, the model still needs to apply significant updates to a small number of complex samples, as reflected by the long tail in the gradient distribution.
	
	An interesting observation for both the Wave network and the Transformer is the opposite trends in the gradient norms of the [CLS] token embedding and the classifier parameters. For the [CLS] token embedding, the gradient is lowest in the first epoch and increases over subsequent epochs, with the peak shifting to the right. This implies that smaller adjustments are made to the [CLS] token embedding early on, but larger updates are necessary later to refine the global semantic representations. In contrast, the classifier parameters exhibit the highest gradient values in the first epoch, which decrease in subsequent epochs, with the peak shifting to the left. This indicates that early in training, larger updates are needed for the classifier to address initial classification errors, but these adjustments diminish as the model approaches convergence. This opposite trend can be explained by the distinct roles of the [CLS] token and the classifier: early on, the [CLS] token provides a rough global representation that requires minimal adjustment, while the classifier must make significant updates to reduce classification errors. As training progresses and the global representation becomes more refined, the [CLS] token demands larger adjustments, while the classifier's updates become smaller, reflecting a shift from early large-scale classifier updates to later fine-tuning stages.
	
	\subsection{Independency Level Among the Dimensions of Embeddings:}
	
	The [CLS] embedding represents global semantics and forms the basis for classification tasks in both wave network and Transformer. This experiment delves into its independence among dimensions to unravel the complex relationship between feature independence and classification accuracy.
	\begin{figure*}[ht]
		\centering
		\includegraphics[width=\textwidth, height=\textheight, keepaspectratio]{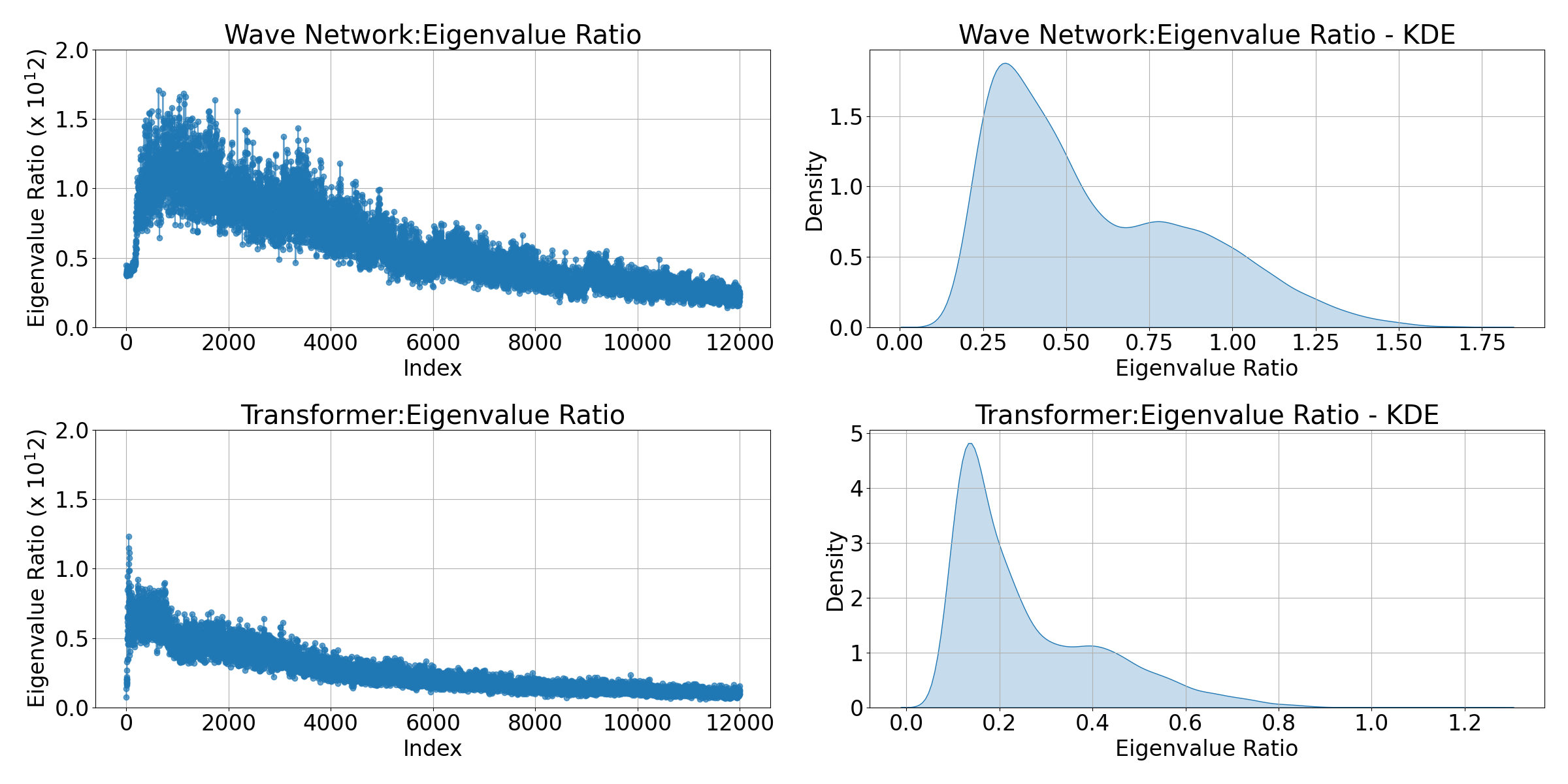}
		\caption{Comparison of dimensional linear independence of first word embedding between wave network (modulation) and the Transformer}
		\label{fig:fullpage1}
	\end{figure*}
	We analyze the correlations between dimensions by examining the covariance matrix of the [CLS] word embedding. Specifically, after training each batch on the AG News dataset, we calculated the minimum and maximum eigenvalues from the [CLS] word embedding matrix and computed their ratio. A significant ratio suggests the presence of a minimal eigenvalue that is close to zero compared to other eigenvalues, indicating that the matrix has reduced influence in the direction corresponding to this eigenvector. This often happens because variations in other directions compensate for the changes in this one. As a result, this direction can be expressed as a linear combination of other dimensions. By analyzing the ratio of the maximum to minimum eigenvalues and its Kernel Density Estimation (KDE), we can gain insights into the dimensional independence and the semantic richness of the word embeddings.
	
	As shown in Figure \ref{fig:fullpage1}, the eigenvalue ratio for the [CLS] token embedding in the Wave network initially rises sharply from around 0.4 to a range of 0.8 to 1.6, indicating strong dimensional correlation in the early stages of training. At this stage, changes in some dimensions are much larger than in others, as the model has not yet learned sufficient information, leading to excessive variance in certain dimensions. As training progresses, the eigenvalue ratio decreases, with the oscillation range shrinking to 0.45-0.65. This suggests that dimensional correlation decreases while independence between dimensions increases. By the fourth epoch, the eigenvalue ratio stabilizes between 0.45 and 0.65, indicating that changes across dimensions become more uniform, with lower correlations and stronger independence.
	
	For the Transformer, the eigenvalue ratio initially fluctuates between 0.1 and 1.25, suggesting significant variation across dimensions and high correlation between them in the early stages of training. However, after about 800 batches, the ratio stabilizes between 0.45 and 0.9 and eventually decreases to 0.4-0.6. This indicates that, as training progresses, the correlation between dimensions decreases, resulting in more uniform changes across dimensions. By the end of training, the ratio converges to 0.08-0.2, indicating that the embedding dimensions have become largely independent, with minimal linear correlation remaining.
	
	In the Wave network, the KDE of the [CLS] token embedding eigenvalue ratio shows that most samples cluster around 0.28, suggesting low correlation and relatively independent information across dimensions for the majority of samples. However, a small platform between 0.65 and 0.8, with a density of 0.75, indicates that some samples exhibit higher dimensional correlations. This suggests that, for these samples, changes in the embedding dimensions are concentrated in fewer directions, resulting in stronger correlations. Although most samples have low eigenvalue ratios, the tail extending to around 1.85 indicates that a few difficult-to-classify samples still exhibit high dimensional correlation, with their dimensional representation concentrated in just a few dimensions.
	
	For the Transformer, the KDE of the [CLS] token embedding eigenvalue ratio shows a peak at 0.13, indicating stronger dimensional independence for most samples. The contributions of changes across dimensions are relatively balanced, suggesting minimal redundancy. A small platform between 0.3 and 0.4, with a density of 1.1, suggests that a few samples exhibit slightly increased correlation. The tail, extending to 1.32, indicates that while most samples show high independence, a few still exhibit some correlation between dimensions. Compared to the Wave network, the Transformer’s KDE distribution is more concentrated, with most samples in a lower range, indicating stronger dimensional independence and less redundancy. The shorter tail confirms that nearly all samples exhibit highly independent embedding dimensions.
	
	The dimensional independence of the [CLS] token embedding in the Transformer is stronger than that in the Wave network. In the Wave network, the magnitude of the [CLS] token embedding represents the global semantics of the input text, calculated as ${input\_G}_k = \sqrt{w_{1,k}^2 + w_{2,k}^2 + \dots + w_{n,k}^2}$. As noted by \citet{osgood1957themeasurement}, semantic dimensions tend to rotate toward each other in complex contexts, forming dependencies. \citet{hollis2016principals} also found that the 300 dimensions representing word meaning are not entirely orthogonal. Our analysis of chance models shows that approximately 280 principal components would be needed to account for 95\% of the variance between dimensions if they were orthogonal. Therefore, when components in one dimension are correlated with others, the magnitude on that dimension reflects information from multiple dimensions, reducing linear independence. Additionally, the phase, representing the angle between the [CLS] token embedding and ${input\_G}_k$, depends on the global semantic characteristics of the input text, further reducing dimensional independence. In contrast, the Transformer’s embedding dimensions are learned through attention allocation between tokens, avoiding the cross-dimensional accumulation effect present in the Wave network.
	
	\subsection{Comparison of Complexity and Parameters}
	Wave network is more efficient and effective in representing tokens and updating representations, significantly reduce the time and space complexity, and parameter quantity. Even its feed-forward and normalization layer must deal with the real and imaginary parts separately in the current PyTorch and TensorFlow versions.
	
	\subsection{Time Complexity}
	
	\noindent\underline{\textit{Transformer:}}\\
	
	\citet{bib5}shows that the complexity of Transformer architecture is:
	\begin{equation}
		\mathcal{O}(n^2 \cdot d)
	\end{equation}
	\noindent\underline{\textit{Wave network:}}
	The time complexity of each step can be analyzed based on batch size, sequence length $n$, and embedding dimension $d$:
	
	\noindent\textbf{1)} \textit \quad {Calculate Source and Target feature:\quad $\mathcal{O}(n \cdot d^2)$}
	
	The source and target features are calculated by two linear transformations from the input feature, which has a shape of [batch size, number of tokens $n$, and dimension of the wave representation of each word $d$]. The linear transformation requires matrix multiplication for each feature vector with a size of d. Then, the computational complexity of a linear layer is $\mathcal{O}(d \times d)$, so for all n nodes, the time complexity is $\mathcal{O}(n \cdot d^2)$.
	
	\noindent\textbf{2)} \textit\quad{Embedding to Complex Vectors:\quad $\mathcal{O}(n \cdot d)$}
	
	This function has a time complexity determined by calculating the magnitude $\mathcal{O}(d)$ and its normalization. Since this operation is performed for each token vector, the time complexity is $\mathcal{O}(n \cdot d)$.
	
	\noindent\textbf{3)} \textit\quad{Complex Addition or Multiplication:\quad $\mathcal{O}(n \cdot d)$}
	
	The complex addition and multiplication here involve element-wise operation, with a shape of [batch size, number of tokens $n$, dimension of the wave representation of each word $d$]. Then, the time complexity is $\mathcal{O}(n \cdot d)$ as the calculation for each complex dimension is independent.

	\noindent\textbf{4)} \textit\quad {Feed-forward:\quad $\mathcal{O}(n \cdot d^2)$}
	
	The feed-forward layer consists of two linear transformations. The first one expands the input dimension from $d$ to $4d$, and the second reduces it back from $4d$ to $d$. Each of these transformations requires matrix multiplication with time complexity $\mathcal{O}(d^2)$ per token. For $n$ tokens, the total time complexity is $\mathcal{O}(n \cdot d^2)$.

	\noindent\textbf{5)} \textit\quad {Normalization:\quad $\mathcal{O}(n \cdot d)$}
	
	The computation for each sample is independent, requiring normalization across the dimension of the wave representation of each word $d$. So for $n$ tokens, the time complexity is $\mathcal{O}(n \cdot d)$.
	
	\noindent\textbf{6)} \textit\quad {Complex Vectors to Embedding:\quad $\mathcal{O}(n \cdot d)$}
	
	Calculating the magnitude and angle has a time complexity of $\mathcal{O}(d)$. Therefore, for $n$ nodes, the time complexity is $\mathcal{O}(n \cdot d)$.
	
	\noindent\textbf{7)} \textit\quad {Summary}
	
	\begin{align}
		\text{Calculate source and target feature: } & \mathcal{O}(n \cdot d^2) \\
		\text{Embedding to complex vectors: } & \mathcal{O}(n \cdot d) \\
		\text{Complex addition/multiplication: } & \mathcal{O}(n \cdot d) \\
		\text{Feed-forward layer: } & \mathcal{O}(n \cdot d^2) \\
		\text{Normalization: } & \mathcal{O}(n \cdot d) \\
		\text{Complex vectors to embedding: } & \mathcal{O}(n \cdot d)
	\end{align}
	
	The time complexity comparison between a single-layer wave network and Transformer is shown in Table 3.
	
	\begin{table*}[h]
		\centering
		\begin{tabular}{cc}
			\hline
			\textbf{Layer Type} & \textbf{Time Complexity} \\
			\hline
			Transformer & $\mathcal{O}(n^2 \cdot d)$ \\
			\hline
			Wave network & $\mathcal{O}(n \cdot d^2)$ \\
			\hline
		\end{tabular}
		\caption{Computational Complexity of Different Layers}
		\label{tab:complexity}
	\end{table*}
	
	\subsection{Space Complexity}
	
	\noindent\underline{\textit{Transformer:}}
	
	\noindent\textbf{1)} \textit\quad{Calculate Q, K, V:\quad $\mathcal{O}(d^2 + n \cdot d)$}
	
	The Query (Q), Key (K), and Value (V) matrices are computed using fully connected linear layers. Each of these layers has a weight matrix of size $d \times d$, resulting in a space complexity of $\mathcal{O}(d^2)$ for each. Additionally, for $n$ tokens, each with an embedding size of $d$, the space required to store these embeddings is $\mathcal{O}(n \cdot d)$. Since Q, K, and V are calculated separately, but each shares this same complexity, the total space complexity for all three remains $\mathcal{O}(d^2 + n \cdot d)$.
	
	\noindent\textbf{2)} \textit\quad{Attention scores and probabilities:\quad $\mathcal{O}(n^2)$}
	
	The attention scores are calculated by performing a dot product between the Query and Key matrices. Given $n$ tokens, this results in an $n \times n$ matrix representing pairwise token relationships. Therefore, the space required to store the attention scores is $\mathcal{O}(n^2)$. The same amount of space is required to store the attention probabilities after applying the softmax function, maintaining the overall space complexity of $\mathcal{O}(n^2)$.
	
	\noindent\textbf{3)} \textit\quad{Context layer calculation:\quad $\mathcal{O}(n \cdot d)$}
	
	The context layer is formed by multiplying the attention probabilities of an $n \times n$ matrix with a same-size Value (V) matrix. This operation results in a new matrix of size $n \times d$. Thus, the space required for storing the context layer is $\mathcal{O}(n \cdot d)$.
	
	\noindent\textbf{4)} \textit\quad{Softmax and head Masking:\quad $\mathcal{O}(n^2)$}
	
	Computing the softmax function for the attention scores and applying any head masking requires storing intermediate results in an $n \times n$ matrix, corresponding to the space used for the attention scores themselves. Thus, the space complexity for this step is also $\mathcal{O}(n^2)$.
	
	\noindent\textbf{5)} \textit\quad{Summary:}
	\begin{align}
		\text{Calculate Q,K,V: } & \quad\mathcal{O}(d^2 + n \cdot d) \\
		\text{Attention scores and probs:: } &\quad \mathcal{O}(n^2) \\
		\text{Context layer calculation: } & \quad\mathcal{O}(n \cdot d) \\
		\text{Softmax and head Masking: } & \quad\mathcal{O}(n^2) \\
		\text{Total video memory: } &\quad \mathcal{O}(n^2 + n \cdot d + d^2) 
	\end{align}
	
	\noindent\underline{\textit{Wave network:}}
	The space complexity of each step can be analyzed based on the batch size, sequence length $n$, and embedding dimension $d$:
	
	\noindent\textbf{1)} \textit\quad{Calculate Source and Target feature:\quad $\mathcal{O}(d^2 + n \cdot d)$}
	
	This complexity arises from fully connected linear layers, each with a weight matrix of size $d \times d$, which requires $\mathcal{O}(d^2)$ space. Additionally, to store the embeddings for $n$ tokens, each of size $d$, we need $\mathcal{O}(n \cdot d)$ space. Hence, the total space complexity is $\mathcal{O}(d^2 + n \cdot d)$.
	
	\noindent\textbf{2)} \textit \quad{Embedding to Complex Vectors:\quad $\mathcal{O}(n \cdot d)$}
	
	Converting embeddings to complex vectors involves storing the magnitude and an angle of a complex number for each of the $n \cdot d$ embedding components, which leads to a space complexity of $\mathcal{O}(n \cdot d)$.

	\noindent\textbf{3)} \textit \quad{Complex Addition or Multiplication: \quad$\mathcal{O}(n \cdot d)$}
	
	For complex addition, each token has a complex embedding of dimension $d$. The operation involves adding embeddings component-wise, which requires storing $n \cdot d$ complex vectors, leading to a space complexity of $\mathcal{O}(n \cdot d)$. Similar to complex addition, complex multiplication is performed element-wise on $n$ tokens, each with an embedding size of $d$. The space required for storing these embeddings remains $\mathcal{O}(n \cdot d)$.

	\noindent\textbf{4)} \textit \quad{Feed-forward: \quad $\mathcal{O}(d^2 + n \cdot d)$}
	
	The feed-forward layers consist of two linear transformations. The first one has a weight matrix of size $d \times (4d)$, and the second has a weight matrix of size $(4d) \times d$. Each layer requires $\mathcal{O}(d^2)$ space. Storing the output of the feed-forward operation for $n$ tokens results in $\mathcal{O}(n \cdot d)$ space. Therefore, the total space complexity is $\mathcal{O}(d^2 + n \cdot d)$.
	
	\noindent\textbf{5)} \textit\quad {Normalization: \quad $\mathcal{O}(d)$}
	
	Layer normalization requires storing a mean and variance for each feature dimension $d$, resulting in a space complexity of $\mathcal{O}(d)$.

	\noindent\textbf{6)} \textit\quad {Complex Vectors to Embedding:\quad $\mathcal{O}(n \cdot d)$}
	
	Converting complex vectors back to embeddings requires storing the real-valued embeddings for all $n$ tokens, each of size $d$, resulting in a space complexity of $\mathcal{O}(n \cdot d)$.
	
	\noindent\textbf{7)} \textit\quad {Summary}
	
	\begin{align}
		\text{Calculate source and target feature: } & \mathcal{O}(d^2 + n \cdot d) \\
		\text{Complex addition/multiplication: } & \mathcal{O}(n \cdot d) \\
		\text{Embedding to complex vectors: } & \mathcal{O}(n \cdot d) \\
		\text{Feed-forward layer: } & \mathcal{O}(d^2 + n \cdot d) \\
		\text{Normalization: } & \mathcal{O}(d) \\
		\text{Complex vectors to embedding: } & \mathcal{O}(n \cdot d)
	\end{align}
	The space complexity comparison between a single-layer wave network and Transformer is shown in Table 4. \\
	\begin{table*}[h]
		\centering
		\begin{tabular}{cc}
			\hline
			\textbf{Layer Type} & \textbf{Space Complexity} \\
			\hline
			Transformer & $\mathcal{O}(n^2 + n \cdot d + d^2)$ \\
			\hline
			Wave network & $\mathcal{O}(d^2 + n \cdot d)$ \\
			\hline
		\end{tabular}
		\caption{Memory consumption of Different Layers}
		\label{tab:complexity}
	\end{table*}

	\subsection{Parameter Estimation}
	
	\noindent\underline{\textit{Bert base:}}
	
	\citet{devlin2019bertpretrainingdeepbidirectional} shows that the amount of parameters of Bert base is 110 million.
	
	\noindent \underline{\textit{Single-layer Wave Network:}}
	
	\noindent\textbf{1)}\textit\quad{Source, Target, and Feed-forward:\quad $(d^2 + d) \cdot 2$}
	
	The source and target transformations are linear layers with dimensions $d \times d$. Each of these layers has $d^2$ parameters for weights and $d$ for biases. Therefore, each transformation layer requires $(d^2 + d)$ parameters. With two such layers, the total parameters are $(d^2 + d) \cdot 2$. The feed-forward network has two linear layers with dimensions $d \times (4d)$ and $(4d) \times d$, requiring $d^2$ parameters for each layer, leading to a similar total parameter count. Hence, the parameters for the feed-forward network are also $(d^2 + d) \cdot 2$.
	
	\noindent\textbf{2)}\textit\quad{Normalization:\quad $d \cdot 2$}
	
	Normalization layers have $d$ parameters for weights and $d$ for biases. Thus, each normalization layer requires $2d$ parameters. Since there are two normalization layers, one for real and one for imaginary parts, the total parameters for normalization are $d \cdot 2$.
	
	\noindent\textbf{3)}\textit\quad{Total Parameters:\quad $(d^2 + d) \cdot 4$}
	
	Combining the parameters from the source, target, feed-forward, and normalization layers gives us a total of $(d^2 + d) \cdot 4$ parameters. 
	
	\noindent\textbf{3)}\textit\quad{When $d = 768$:}
	
	Substituting $d = 768$ into the total parameters formula, we get:
	\begin{equation}
		(768^2 + 768) \cdot 4 = 2,365,184.
	\end{equation}
	The comparison of Bert base and our single-layer wave network is shown in Table 5.
	\begin{table*}[ht]
		\centering
		\begin{tabular}{cc}
			\hline
			\textbf{Layer Type} & \textbf{Parameters} \\
			\hline
			Bert base & 110 million \\
			\hline
			Wave network & 2.37 million \\
			\hline
		\end{tabular}
		\caption{Parameter comparison}
		\label{tab:complexity}
	\end{table*}
	
	\section{Discussion}\label{sec3}
	We introduced Token2Wave, a novel representation method that captures both global and local text semantics. Token2Wave offers significant potential to transform fields that rely heavily on the NLP by delivering expected performance with minimal computational hardware and reduced processing time. This efficiency suggests that many devices with limited hardware resources could perform advanced NLP tasks, including personalized fine-tuning and reasoning, without relying on extensive infrastructure. These benefits could profoundly improve existing business environments. For instance, Token2Wave could enable medical devices with limited computing resources to incorporate natural language understanding for supporting remote diagnosis and real-time health monitoring or provide personalized psychological support and advice based on user sentiment on a mobile device.

	\bibliography{reference}

\end{document}